%% file: main.tex
\documentclass[runningheads]{llncs}
\usepackage[T1]{fontenc}
\usepackage{graphicx}
\usepackage{xcolor}

\begin{document}
\title{Question Personalization in an Intelligent Tutoring System}
\titlerunning{Question Personalization in an ITS}
\author{Sabina Elkins\inst{1, 2} \and Robert Belfer \inst{2} \and Ekaterina Kochmar \inst{2, 3} \and Iulian Serban \inst{2} \and Jackie C.K. Cheung \inst{1, 4}}
\authorrunning{S. Elkins et al.}
\institute{McGill University \& MILA (Quebec Artificial Intelligence Institute)
\and Korbit Technologies Inc. \and University of Bath \and Canada CIFAR AI Chair}

\maketitle 
\begin{abstract}
This paper investigates personalization in the field of intelligent tutoring systems (ITS).
We hypothesize that personalization in the way questions are asked improves student learning outcomes. 
Previous work on dialogue-based ITS personalization has yet to address question phrasing.
We show that generating versions of the questions suitable for students at different levels of subject proficiency improves student learning gains, using variants written by a domain expert and an experimental A/B test. 
This insight demonstrates that the linguistic realization of questions in an ITS affects the learning outcomes for students.

\end{abstract}

\section{Introduction}
\input{sections/introduction}

\section{Methodology}\label{methods}
\input{sections/methodology}

\section{Results and Analysis}
\input{sections/results_v2}

\section{Conclusion}
\input{sections/future}

\subsubsection{Acknowledgements} 
We'd like to thank Korbit for hosting our experiment on their platform, and Mitacs for their grant to support this project. We are grateful to the anonymous reviewers for their valuable feedback.

\input{biblio}
\end{document}

%% file: sections/introduction.tex
Intelligent tutoring systems (ITS) are AI systems capable of automating teaching. They have the potential to provide accessible and highly scalable education to students around the world \cite{kulik_2016}.
Previous studies suggest that students learn significantly better in one-on-one tutoring settings than in classroom settings \cite{bausell_1972}. Personalization can be addressed in an AI-driven, dialogue-based ITSs, and can have significant impact on the learning process \cite{kochmar_2020}. 
This has been explored in different ways, including dialogue feedback and question selection \cite{st2022new}. To the best of our knowledge, personalization in question phrasing has not been explored.

Students benefit from being asked questions tailored to their level of subject expertise and their needs during in-person tutoring sessions \cite{ashton_1988,hrastinski_2019}. We hypothesize that the same effect can be achieved when questions are adapted to the students' levels of expertise and their needs in an ITS. 
To test this, we integrate question variants created by a human domain expert onto the Korbit Technologies Inc. platform and run an A/B test. Korbi's AI tutor, \texttt{Korbi}, is a dialogue-based ITS, which teaches students by providing them with video lectures and interactive problem solving exercises, selected for each student using ML and NLP techniques \cite{serban2020large}. The main contribution of this paper is the demonstration that question personalization in an ITS leads to improvements in learning gains.

%% file: sections/methodology.tex
Students interact with \texttt{Korbi} through short answer questions and written responses. To assess if the phrasing of these questions can impact learning gains, we first create a set of questions and variants that reformulate the original idea. The variants were created by a human expert from existing questions on the Korbit platform. They were designed to reflect three levels of difficulty: \textit{beginner}, \textit{intermediate}, and \textit{advanced}, as per common practice in education.
We assume that less detailed questions are harder (as the student must have more background knowledge to understand and answer) and more elaborate ones are easier (as they `hint' at the answer with extra information) \cite{taylor_1962}. 
In our data, each question has three variants at different levels of proficiency. Questions were made easier by adding elaborations and synonym replacement, and more difficult by removing non-essential explanations and synonym replacement. As Table \ref{tab1} shows, the beginner variants are longer and the advanced ones are more concise.

\vspace{-6mm} \begin{figure} \begin{center}
\includegraphics[width=0.7\textwidth]{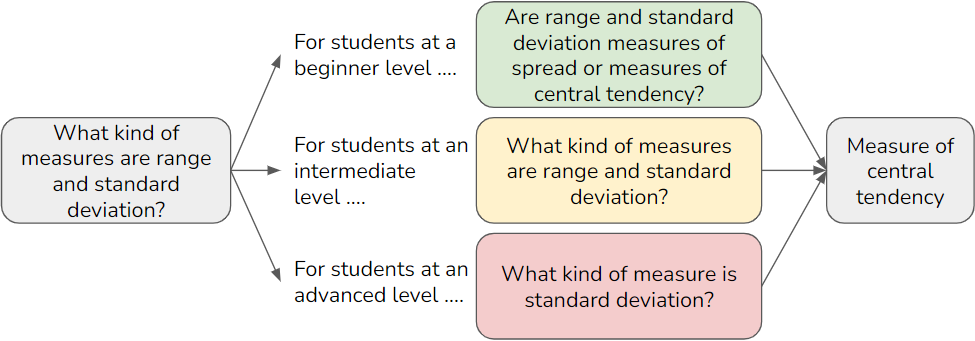} \vspace{-3mm}
\caption{An question being adapted to different levels while retaining the same answer.} \label{fig1}
\end{center} \end{figure} \vspace{-8mm}

\noindent The variants were given to three human experts (data scientists with at least an MSc in a related field) who rated them on three scales from $0$ to $5$, representing \textit{difficulty} (i.e., the relative complexity of the question as compared to the others), \textit{fluency} (i.e., spelling and grammar), and \textit{meaning preservation} (i.e., if the meaning of the original question is preserved). The mean results of their ratings can be seen in Table \ref{tab1}. The \textit{fluency} and \textit{meaning preservation} metrics are consistently high and the \textit{difficulty} metric increases with the assigned levels. 

\begin{table} \begin{center}\vspace*{-\baselineskip}
    \caption{Mean variant scores from human experts, and average word counts by level.}\label{tab1} \vspace*{-0.5\baselineskip}
    \begin{tabular}{| l | c | c | c || c |}
    \hline
    Level & Difficulty &  Fluency & Meaning Preservation & Mean Word Count \\
    \hline
    Beginner & 1.689 $\pm 0.635 $ & 4.600 $\pm 0.471$ & 4.789 $\pm 0.451$ & 39.800\\
    Intermediate & 2.667 $\pm 0.689$ & 4.683 $\pm 0.481$ & 4.839 $\pm 0.406$ & 33.533\\
    Advanced & 3.939 $\pm 1.269$ & 4.544 $\pm 0.661$ & 4.717 $\pm 0.516$ & 27.433\\
    \hline
    \end{tabular}
\end{center} \end{table} \vspace*{-2\baselineskip}

\noindent The next task is to select an appropriate question variant for each student at each step in the dialogue. Through Korbit, we have anonymized access to student history. We isolated $2{,}137$ students' interactions with the platform. Each student's history consists of the exercises they encountered and their attempts to solve them. Each exercise encountered was included as a point in our dataset, for a total of $13{,}504$ exercises given to students. Using this, it is possible to calculate a set of features indicative of a student's level, and subsequently build a logistic regression model to predict if a student will succeed on the next exercise.

The original feature set consisted of 7 features, including overall success rate, improvement (i.e., changes in success rate), skip rates, and others. From this set, two features were selected based on their contribution to the best model in the preliminary experiments: (1) \textit{topic success feature} is a numerical feature in $[0,1]$ that shows the eventual success rate per all exercises previously attempted in a given topic, and (2) \textit{topic skip feature} is a numerical feature in $[0,1]$ that is the skip rate per all exercises previously attempted in a given topic. A topic on \texttt{Korbi} is a broad category of material, such as `Probability'. Using these features the model is able to predict next exercise success with an accuracy of $80\%$.

The variant assignment model calculates the features when a student gets a new exercise, and generates a probability of success with the regression model. Students are assigned variants based on the percentile range that their probability of success falls into (calculated from the predictions across the entire data set). Students in the $0^{th}$ to $33^{rd}$ percentiles get beginner variants, in the $33^{rd}$ to $66^{th}$ percentiles get intermediate variants, and the rest get advanced ones.

%% file: sections/results_v2.tex
To test our claims, we put the variants and assignment model described in Section \ref{methods} on the \texttt{Korbi} platform. Our A/B test ran over 2 months, collecting data from over 400 students at varied skill levels. Student attempts were divided into three groups. The \textit{expected} variant group received the variant which matched their assignment model score. The \textit{non-expected} variant group received a variant which did not match their score from the assignment model (e.g., beginner question for an advanced student). The \textit{control} group students received the original variant (i.e., that which was already on the platform before this experiment).

\begin{table} \begin{center} \vspace*{-\baselineskip}
    \caption{Test results. Metrics marked with * 
    are statistically significant at the $\alpha = 0.05$ level by a Student's $t$-test.} \label{tab2}
    \begin{tabular}{| l | c | c | c | c |}
    \hline
    Experiment Group & Solution Acceptance* & Ultimate Failure Rate* & Skip Rate & n \\
    \hline
    Expected & $0.626 \pm 0.069$ & $0.163 \pm 0.053$ & $0.105 \pm 0.044$ & 190\\
    Non-Expected & $0.468 \pm 0.083$ & $0.295 \pm 0.076$ & $0.144 \pm 0.058$ & 139\\
    Control & $0.596 \pm 0.081$ & $0.191 \pm 0.065$ & $0.121 \pm 0.054$ & 141\\
    \hline
    \end{tabular}
\end{center} \end{table} \vspace*{-2\baselineskip}

\noindent \textit{Solution acceptance} is the proportion of success per exercise attempts. 
However, succeeding on exercises does not equate to learning. Students should be challenged within their zone of proximal development \cite{cazden_1979} but eventually obtain the right answer, so we aim to minimize the \textit{ultimate failure rate} as opposed to simply maximizing attempt success. 
This metric is the proportion of failure out of all exercises seen by students. Unlike \textit{solution acceptance} which shows the success rate per attempt, \textit{ultimate failure rate} shows the fail rate per exercise.
\textit{Skip rate} is indicative of a student's engagement. Intuitively, the more they skip, the less they engage with the content.
All three of these metrics show the \textit{expected} group performing the best, followed by \textit{control} and finally \textit{non-expected}. For \textit{solution acceptance} and \textit{ultimate failure rate}, the difference between \textit{expected} and \textit{non-expected} groups is statistically significant at $\alpha = 0.05$ by a Student's $t$-test.

The difference between the \textit{expected} and \textit{control} groups is smaller than the difference between the \textit{expected} and \textit{non-expected} groups. 
This can be attributed to the fact that the original questions were refined through several rounds of review by domain experts when they were created for \texttt{Korbi} platform, whereas the variants only were reviewed once.
Additionally, the \textit{control} group's exercises are always intermediate or advanced, while the strongest result is seen with beginners. Isolating the students who score for beginner variants only, we see a $19\%$ relative reduction in \textit{ultimate failure} when comparing the \textit{expected} to the \textit{control} group, which demonstrates a bigger impact for beginners.
Additionally, the same comparison shows a $30\%$ relative reduction in the \textit{skip rate}, suggesting that the beginners are more engaged when dealing with beginner variants.


%% file: sections/future.tex
We see a clear improvement in the success of students in the \textit{expected} group. This confirms our hypothesis that providing question variants suited to student's level improves their learning gains. These variants are more useful for beginner students who need more assistance, which is an encouraging and intuitive result.
The future of this work is in automating the creation question variants for scalability, and creating a more sophisticated variant assignment approach.

%% file: biblio.tex
\bibliographystyle{splncs04}